\documentclass[10pt,twocolumn,letterpaper]{article}

\usepackage{cvpr}
\usepackage{times}
\usepackage{epsfig}
\usepackage{graphicx}
\usepackage{amssymb}
\usepackage{amsmath,bm}
\usepackage{multirow}
\usepackage{booktabs}
\usepackage[table]{xcolor}
\definecolor{mygray}{rgb}{0.6902, 0.76863, 0.87059}

\usepackage[pagebackref=true,breaklinks=true,letterpaper=true,colorlinks,bookmarks=false]{hyperref}

\cvprfinalcopy 

\def\cvprPaperID{1570} 

\ifcvprfinal\fi
\begin{document}

\title{See More, Know More: Unsupervised Video Object Segmentation with \\Co-Attention Siamese Networks}
\author{Xiankai Lu$^{1}\thanks{The first two authors contribute equally to this work.}$~,\hspace{1pt} Wenguan Wang$^{1*}$,\hspace{1pt} Chao Ma$^{2}$,\hspace{1pt}   Jianbing Shen$^{1}$\thanks{Corresponding author: \textit{Jianbing Shen}. }, \hspace{2pt}  Ling Shao$^{1}$,\hspace{1pt}  Fatih Porikli$^{3}$  \\
	\small{$^1$} \small Inception Institute of Artificial Intelligence, UAE \hspace{0pt} \\
	\small{$^2$} \small MoE Key Lab of Artificial Intelligence, AI Institute, Shanghai Jiao Tong University, China \hspace{0pt} \\
	\small{$^3$} \small Australian National University, Australia \\
	{\tt\small carrierlxk@gmail.com} \hspace{2pt}
	{\tt\small wenguanwang.ai@gmail.com} \hspace{2pt} {\tt\small chaoma@sjtu.edu.cn} \\
		{\tt\small shenjianbingcg@gmail.com} \hspace{2pt} {\tt\small ling.shao@ieee.org} \hspace{2pt}
		{\tt\small fatih.porikli@anu.edu.au} \\
		{\tt\small \url{https://github.com/carrierlxk/COSNet}}
	%
}

\maketitle
\thispagestyle{empty}




\begin{abstract}

We introduce a novel network, called CO-attention Siamese Network (COSNet), to address the unsupervised video object segmentation task from a holistic view. We emphasize the importance of inherent correlation among video frames and incorporate a global co-attention mechanism to improve further the state-of-the-art deep learning based solutions that primarily focus on learning discriminative foreground representations over appearance and motion in short-term temporal segments. The co-attention layers in our network provide efficient and competent stages for capturing global correlations and scene context by jointly computing and appending co-attention responses into a joint feature space. We train COSNet with pairs of video frames, which naturally augments training data and allows increased learning capacity. During the segmentation stage, the co-attention model encodes useful information by processing multiple reference frames together, which is leveraged to infer the frequently reappearing and salient foreground objects better. We propose a unified and end-to-end trainable framework where different co-attention variants can be derived for mining the rich context within videos.
Our extensive experiments over three large benchmarks manifest that COSNet outperforms the current alternatives by a large margin. 

\end{abstract}
\vspace*{-8pt}

\section{Introduction}\label{sec:intro}
Unsupervised video object segmentation (UVOS) aims to automatically separate primary foreground object(s) from their background in a video. Since UVOS does not require manual interaction, it has significant value in both academic and applied fields, especially in this  era of information-explosion. However, due to the lack of prior knowledge about the primary object(s), in addition to the typical challenges for semi-supervised video object segmentation (\eg, object deformation, occlusion, and background clutters), UVOS suffers from another difficult problem, \ie, how to correctly distinguish the primary objects from a complex and diverse background.

\begin{figure}
	\centering
	\includegraphics[width=.48\textwidth]{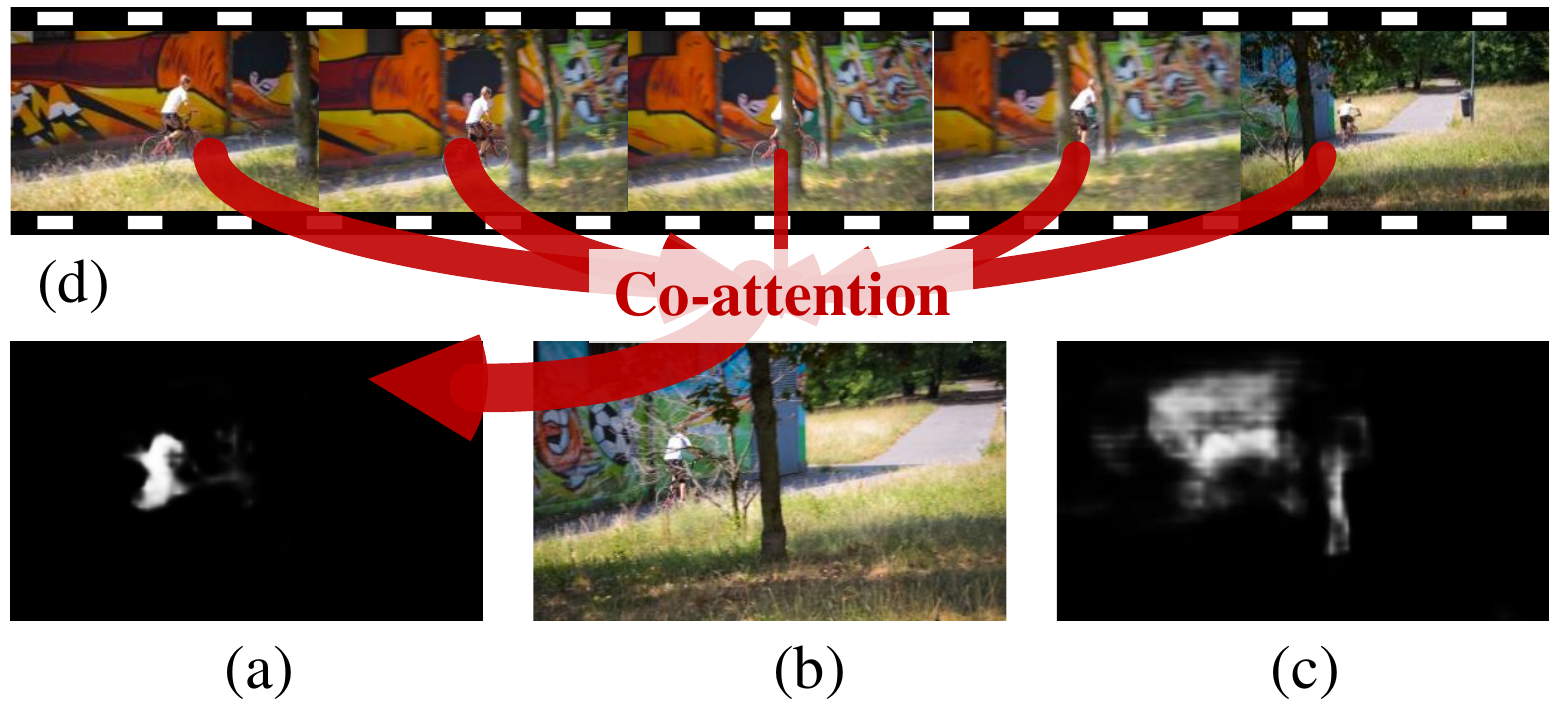}
	\caption{Illustration of our intuition. Given an input frame (b), our method leverages information from multiple reference frames (d) to better determine the foreground object (a), through a co-attention mechanism. (c) An inferior result without co-attention. 
	}
	\label{fig:demo}
	\vspace{-12pt}
\end{figure}

We argue that the primary objects in UVOS settings should be the most (\textbf{i}) distinguishable in an individual frame (locally salient), and (\textbf{ii}) frequently appearing throughout the video sequence (globally consistent). These two properties are essential for determining the primary objects. For instance, by only glimpsing a short video clip as illustrated in Fig.~\ref{fig:demo}(b), it is hard to determine the primary objects. Instead, if we view the entire video (or a sufficiently long sequence) as in Fig.~\ref{fig:demo}(d), the foreground can be easily discovered. Although primary objects tend to be highly correlated at a macro level (entire video), they often exhibit different appearances at a micro level (shorter video clips) due to articulated body motions, occlusions, out-of-view movements, camera movements, and environment variations. Clearly, micro level variations are the major sources of challenges in video segmentation. Thus, it is desirable to take advantage of the global consistency property and leverage the information from other frames.

By considering UVOS from a global perspective, we can help to locate primary objects and alleviate the local ambiguities. This notion also motivated the earlier heuristic models for UVOS~\cite{DBLP:conf/bmvc/FaktorI14}, yet it is largely ignored by current deep learning based models.


Current deep learning based UVOS models typically focus on the intra-frame discrimination property of primary objects in appearance or motion, while ignoring the valuable global-occurrence consistency across multiple frames. These methods compute optical flows across a few consecutive frames~\cite{DBLP:conf/iccv/TokmakovAS17,jain2017fusionseg,cheng2017segflow,Li_2018_CVPR,Li_2018_ECCV1}, which is limited to a local receptive window in the temporal domain. Although recurrent neural networks (RNNs)~\cite{Song_2018_ECCV} are introduced to memorize previous frames, this sequential processing strategy may fail to explicitly explore the rich relations between different frames, hence does not attain a global perspective.


With these insights, we reformulate the UVOS task as a co-attention procedure and propose a novel CO-attention Siamese Network (COSNet) to model UVOS from a global perspective.
Specifically, during the training phase, COSNet takes a pair of frames from the same video as input and learns to capture their rich correlations. This is achieved by a differentiable, gated co-attention mechanism, which enables the network to attend more to the correlated, informative regions, and produce further discriminative foreground features. For a testing frame (Fig.~\ref{fig:demo}(b)), COSNet is able to produce more accurate results (Fig.~\ref{fig:demo}(a)) from a global view, \ie, utilize the correlations between the testing frame and multiple reference frames. Fig.~\ref{fig:demo}(c) shows the inferior result when considering only the information from the testing frame (Fig.~\ref{fig:demo}(b)).


Another advantage of our COSNet is that it is remarkably efficient for augmenting training data, as it allows using a large number of arbitrary frame pairs within the same video. 
Additionally, as we explicitly model the relations between video frames, the proposed model does not need to compute optical flow, which is time-consuming and computationally expensive. Finally, the COSNet offers a unified, end-to-end trainable framework that efficiently mines rich contextual information within video sequences. We implement different co-attention mechanisms such as vanilla co-attention, symmetric co-attention, and channel-wise co-attention, which offers a more insightful glimpse into the task of UVOS. We quantitatively demonstrate that our co-attention mechanism is able to bring large improvement in performance, which confirms its effectiveness and the value of global information for UVOS. The proposed COSNet shows superior performance over the current state-of-the-art methods across three popular benchmarks: DAVIS16~\cite{perazzi2016benchmark}, FBMS~\cite{DBLP:journals/pami/OchsMB14} and Youtube-Objects~\cite{DBLP:conf/cvpr/PrestLCSF12}. 



\section{Related Work}
We start by providing an overview of representative work on video object segmentation (\S\ref{sec:vos}), followed by a brief overview of differentiable neural attention (\S\ref{sec:amnn}).

\subsection{Video Object Segmentation}\label{sec:vos}
According to its supervision type, video object segmentation can be broadly categorized into \textit{unsupervised} (UVOS) and \textit{semi-supervised} video object segmentation. In this paper, we focus on the UVOS task, which extracts primary object(s) without manual annotation.

Early \textbf{UVOS models} typically analyzed long-term motion information (trajectories)~\cite{Brox2010,DBLP:conf/iccv/OchsB11,DBLP:conf/cvpr/FragkiadakiZS12,DBLP:conf/iccv/PapazoglouF13,DBLP:conf/iccv/KeuperAB15,DBLP:journals/pami/OchsMB14},  
 leveraged object proposals~\cite{lee2011key,ma2012,zhang2013,DBLP:conf/cvpr/KohK17,fu2014object,Koh_2018_ECCV} or utilized saliency information~\cite{DBLP:conf/cvpr/WangSP15,DBLP:conf/bmvc/FaktorI14,tsai2016semantic,hu2018unsupervised}, 
to infer the target. Later, inspired by the success of deep learning, several methods~\cite{fragkiadaki2015learning,DBLP:conf/cvpr/Tsai0B16,pathak2017learning} began to approach UVOS using deep learning features. These were typically limited due to their lack of end-to-end learning ability~\cite{DBLP:conf/cvpr/Tsai0B16} and use of heavyweight fully-connected network architectures~\cite{fragkiadaki2015learning,pathak2017learning}. Recently, more research efforts have focused on the fully convolutional neural network based UVOS models. For example, Tokmakov \etal~\cite{DBLP:conf/cvpr/TokmakovAS17} proposed to separate independent object and camera motion using a learnable motion pattern network~\cite{DBLP:conf/cvpr/TokmakovAS17}. Li \etal~learned an instance embedding network~\cite{Li_2018_CVPR} from static images to better locate the object(s), and later they combined motion-based bilateral networks 
for identifying the background~\cite{Li_2018_ECCV1}.  Two-stream fully convolution networks are also a popular choice~\cite{cheng2017segflow,jain2017fusionseg,DBLP:conf/iccv/TokmakovAS17,Li_2018_CVPR} to fuse motion and appearance information together for object inference.
An alternative way to segment an object is through video salient object detection~\cite{Song_2018_ECCV}. This method fine-tunes the pre-trained semantic segmentation network for extracting spatial saliency features, then trains ConvLSTM  to capture temporal dynamics. 

These deep UVOS models generally achieved promising results, which demonstrates well the advantages of applying neural networks to this task. However, they only consider the sequential nature of UVOS and short-term temporal information, lacking a global view and comprehensive use of the rich, inherent correlation information within videos.
\begin{figure*}[t]
	\centering
	\includegraphics[width=.9\textwidth]{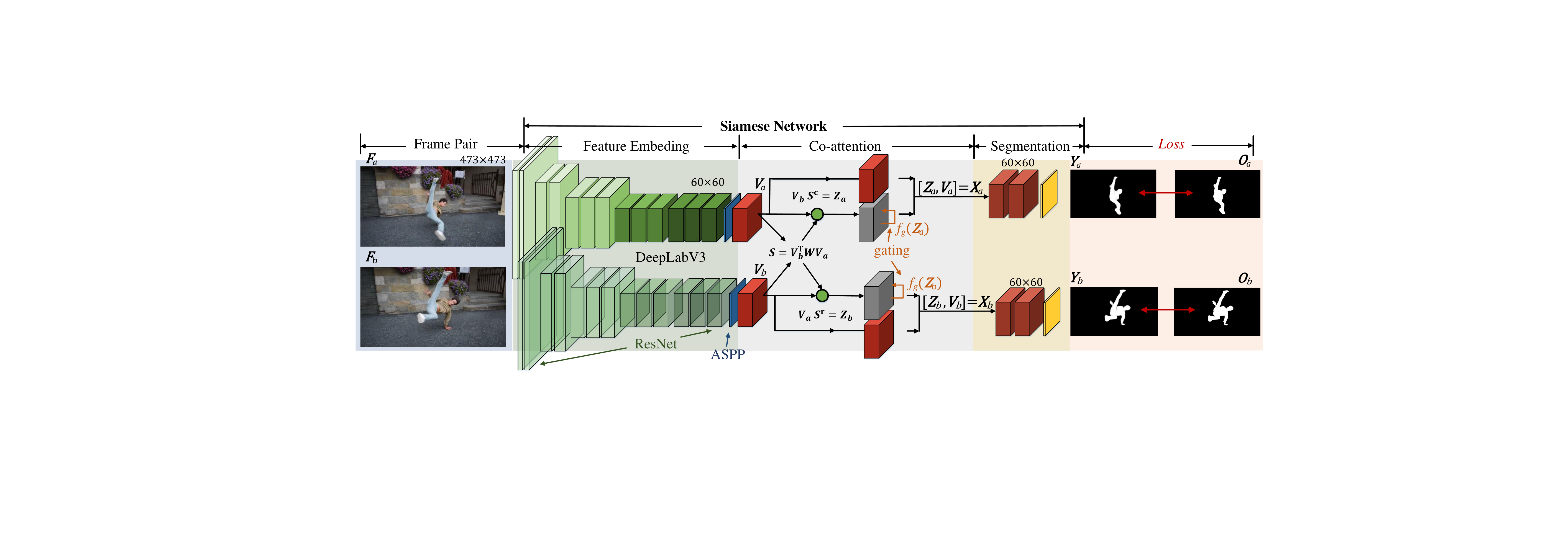}
	\caption{\small Overview of COSNet in the training phase. 
A pair of frames $\{\mathbf{F}_{\!a}, \mathbf{F}_{\!b}\}$ is fed into a feature embedding module to obtain the feature representations $\{\mathbf{V}_{\!a}$, $\mathbf{V}_{\!b}\}$. Then, the co-attention module computes the attention summaries that encode the correlations between $\mathbf{V}_{\!a}$ and $\mathbf{V}_{\!b}$. Finally, $\mathbf{Z}$ and $\mathbf{V}$ are concatenated and handed over to a segmentation module to produce segmentation predictions.  }
\vspace*{-4pt}
	\label{fig:framework}
\end{figure*}

For \textbf{SVOS methods}, the target object(s) is provided in the first frame and tracked automatically~\cite{DBLP:conf/cvpr/WangSP15,cheng2018fast,DBLP:conf/cvpr/CaellesMPLCG17,yang2018efficient,Bao_2018_CVPR,DBLP:conf/iccv/YoonRKLSK17,xiao2018monet,zuo2018learning} or interactively by users~\cite{Bai2009} 
  in the subsequent frames. Numerous algorithms were proposed based on graphical models~\cite{DBLP:conf/cvpr/Tsai0B16}, object proposals~\cite{DBLP:conf/iccv/PerazziWGS15}, super-trajectories~\cite{wang2017}, \etc.  Recently, deep learning based methods achieved promising results. Some algorithms treated video object segmentation as a static segmentation task without using any temporal information~\cite{DBLP:conf/cvpr/PerazziKBSS17}, built a deep one-shot learning framework~\cite{DBLP:conf/cvpr/CaellesMPLCG17,wang2015robust}, or used a mask-propagation network~\cite{DBLP:conf/cvpr/JampaniGG17}. In addition, both object tracking~\cite{DAVIS2017-2nd,cheng2018fast,Ci_2018_ECCV,Lu_2018_ECCV} and person re-identification~\cite{Li_2018_ECCV,DBLP:conf/eccv/YanNSMYY16} have been fused into SVOS task to handle  deformation and occlusion issues. Hu \etal.~\cite{Hu_2018_ECCV} proposed a Siamese network based SVOS model. Compared with our COSNet, the differences are distinct, rather than their dissimilar supervision manners. First, since~\cite{Hu_2018_ECCV} was proposed based on image matching strategy, they used a Siamese network to propagate the first-frame annotation to the subsequent frames. Our method substantially differs in that we learn the Siamese network to capture rich and global correspondences within videos to further assist automatic primary object discovery and segmentation. Second, we provide the first approach that uses a co-attention scheme to facilitate correspondence learning for video object segmentation.

\subsection{Attention Mechanisms in Neural Networks}\label{sec:amnn}
Differentiable attentions, which are inspired by human perception~\cite{DBLP:journals/neco/DenilBLF12,wang2018deep}, have been widely studied in deep neural networks~\cite{jetley2018learn,DBLP:conf/nips/VaswaniSPUJGKP17,mnih2014recurrent,DBLP:conf/nips/JaderbergSZK15,DBLP:conf/cvpr/WangJQYLZWT17, NonLocal2018,Fang_2018_ECCV}. With end-to-end training, neural attention allows networks to selectively pay attention to a subset of inputs. For example,
Chu \etal.~\cite{DBLP:conf/cvpr/ChuYOMYW17} exploited multi-context attention for human pose estimation. In~\cite{DBLP:conf/cvpr/ChenZXNSLC17}, spatial and channel-wise attention were proposed to dynamically select an  image part for captioning.

More recently, co-attention mechanisms have been studied in vision-and-language tasks, such as visual question answering~\cite{DBLP:conf/nips/LuYBP16,DBLP:journals/corr/XiongZS16,Wu_2018_CVPR,Nguyen_2018_CVPR} and visual dialogue~\cite{Wu_2018_CVPR}. In these works, co-attention mechanisms were used to mine the underlying correlations between different modalities. For example, Lu \etal~\cite{DBLP:conf/nips/LuYBP16} created a model that jointly performs question-guided visual attention and image-guided question attention. In this way, the learned model can selectively focus on image regions and segments of documents. Our co-attention model is inspired by these works, but it is used to capture the coherence across different frames with a more elegant network architecture.
\section{Proposed Algorithm}
Our COSNet formulates UVOS as a co-attention procedure. A co-attention module learns to explicitly encode correlations between video frames. This enables COSNet to attend to the frequently coherent regions, thus further helping to discover the foreground object(s) and produce reasonable UVOS results. Specifically, during training, co-attention procedure can be decomposed into the correlation learning between any frame pairs from the same video (see Fig.~\ref{fig:framework}). 
During testing, COSNet infers the primary target with a global view, \ie, takes advantage of the co-attention information between the testing frame and multiple reference frames. We will elaborate the co-attention mechanisms in COSNet in \S\ref{sec:method}, and detail the whole architecture of COSNet in \S\ref{sec:architecture}. In \S\ref{sec:dna}, we will provide more implementation details.
\subsection{Co-attention Mechanisms in COSNet}
\label{sec:method}
\begin{figure}[t]
	\centering
	\includegraphics[width=.49\textwidth]{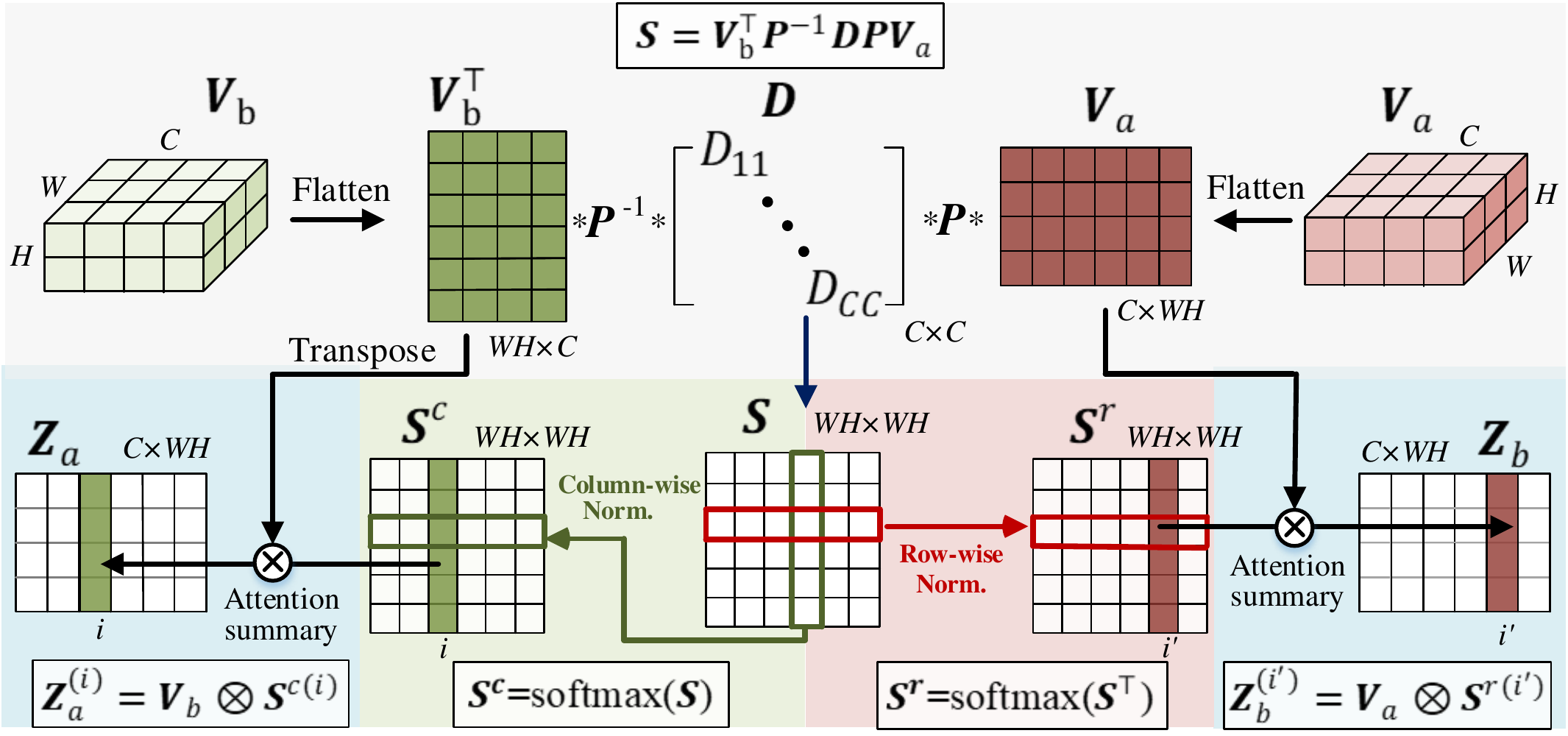}
	\caption{ Illustration of our co-attention operation.  }
	\label{fig:matrix}
	\vspace{-10pt}
\end{figure}

\noindent\textbf{Vanilla co-attention.}
As shown in Fig.~\ref{fig:framework}, given two video frames $\mathbf{F}_a$ and $\mathbf{F}_b$ from the same video, $\mathbf{V}_a \!\in\! \mathbb{R}^{W\!\times \!H\!\times\! C}$ and $\mathbf{V}_b \!\in\! \mathbb{R}^{W\!\times\! H\!\times\! C}$ denote the corresponding feature representations from a feature embedding network. $\mathbf{V}_a$ and $\mathbf{V}_b$ are 3D-tensors with the width $W$, height $H$ and  $C$ channels. We leverage the co-attention mechanism~\cite{DBLP:journals/corr/XiongZS16,DBLP:conf/nips/LuYBP16} to mine the correlations between $\mathbf{F}_a$ and $\mathbf{F}_b$ in their feature embedding space. More specifically, we first compute the affinity matrix $\mathbf{S}$ between $\mathbf{V}_a$ and $\mathbf{V}_b$: 
\begin{equation}\small
\mathbf{S} = {\mathbf{V}^\top_b} \mathbf{W} {\mathbf{V}_a} \in \mathbb{R}^{(W\!H)\times (W\!H)},
\label{co-attention}
\end{equation}
where  $\mathbf{W} \!\in\! \mathbb{R}^{C\times C}$ is a weight matrix. Here $\mathbf{V}_{\!a\!} \!\in\!\! \mathbb{R}^{C\times {(W\!H)}\!\!}$ and $\mathbf{V}_b \!\in\! \mathbb{R}^{C\times{(W\!H)}\!}$  are flattened into matrix representations.  Each column ${\mathbf{V}^{(i)}_{\!a}}$  in $\mathbf{V}_{\!a}$ represents the feature vector at position $i \!\in\! \{1,...,W\!H\}$ with $C$ dimensions. As a result, each entry of $\mathbf{S}$ reflects the similarity between each row of $\mathbf{V}^\top_{\!b}$ and each column of $\mathbf{V}_{\!a}$. Since the weight matrix  $\mathbf{W}$ is a square matrix, the diagonalization of $\mathbf{W}$ can be represented as follows:
\begin{equation}\small
\mathbf{W} = \mathbf{P}^{-1} \mathbf{D} \mathbf{P},
\label{diag}
\end{equation}
where $\mathbf{P}$ is an invertible matrix and $\mathbf{D}$ is a diagonal matrix. Then, as shown in the gray area in Fig.~\ref{fig:matrix}, Eq.~\ref{co-attention} can be re-written as:
\begin{equation}\small
\mathbf{S} = {\mathbf{V}^\top_b}\mathbf{P}^{-1} \mathbf{D} \mathbf{P} {\mathbf{V}_a}. 
\label{co-attention1}
\end{equation}

Through the \textit{vanilla co-attention} in Eq.~\ref{fig:matrix}, the feature representation of each frame first undergoes linear transformations, and then calculates the distance between any locations of themselves.

\noindent\textbf{Symmetric co-attention.}
If we further constrain the weight matrix to be a symmetric matrix, the project matrix $\mathbf{P}$ becomes an orthogonal matrix: $\mathbf{P}^\top\mathbf{P} \!=\! \mathit{I}$, where $\mathit{I}$ is a $C$$\times$$C$ identity matrix. A~\textit{symmetric co-attention} can be derived from Eq.~\ref{co-attention1}:
\begin{equation}\small
\begin{aligned}
\mathbf{S} = {\mathbf{V}^\top_b} \mathbf{P}^{\top} \mathbf{D} \mathbf{P} {\mathbf{V}_a}
= \left( \mathbf{P}\mathbf{V}_b\right)^{\top}\mathbf{D} \mathbf{P} {\mathbf{V}_a}.
\end{aligned}
\label{orthonal}
\end{equation}
Eq.~\ref{orthonal} indicates that we project the feature embeddings $\mathbf{V}_a$  and $\mathbf{V}_b$ into an orthogonal common space  and maintain their norm of $\mathbf{V}_a$  and $\mathbf{V}_b$. This  property has proved valuable for  eliminating the correlation between different channels (\ie., $C$- dimension)~\cite{DBLP:conf/iccv/SunZDW17} and  improving the network's generalization ability~\cite{brock2016neural,rodriguez2016regularizing}.

\noindent\textbf{Channel-wise co-attention.}
Furthermore, the project matrix $\mathbf{P}$ can be simplified into an identity matrix $\mathit{I}$ (\ie., without space transformation), and then the weight matrix $\mathbf{W}$ becomes a diagonal matrix. In this case, $\mathbf{W}$ (\ie., $\mathbf{D}$)  can be further diagonalized into two  diagonal matrices $\mathbf{D}_a$ and $\mathbf{D}_b$.  Thus, Eq.~\ref{co-attention1} can be re-written as \textit{channel-wise co-attention}:
\begin{equation}\small
\begin{aligned}
\mathbf{S} = {\mathbf{V}^\top_{\!b}} \mathit{I}^{-1} \mathbf{D} \mathit{I} {\mathbf{V}_{\!a}}
= {\mathbf{V}^\top_{\!b}} {\mathbf{D}^\top_a} {\mathbf{D}_b} {\mathbf{V}_{\!a}}
=\left(\mathbf{D}_a \mathbf{V}_b\right)^\top {\mathbf{D}_b \mathbf{V}_{\!a}}.
\end{aligned}
\label{channel-attention1}
\end{equation}
This operation is equal to applying a channel-wise weight to $\mathbf{V}_{\!a}$ and $\mathbf{V}_{\!b}$  before computing the similarity. This helps to alleviate channel-wise redundancy, which shares a similar spirit to Squeeze-and-Excitation mechanism~\cite{DBLP:conf/cvpr/ChenZXNSLC17,hu2018senet}.
During ablation studies (\S\ref{sec:ablation}), we perform detailed experiments to assess the effect of the different co-attention mechanisms, \ie., vanilla co-attention (Eq.~\ref{co-attention1}), symmetric co-attention (Eq.~\ref{orthonal}) and channel-wise co-attention (Eq.~\ref{channel-attention1}).

\begin{figure*}[t]
	\centering
	\includegraphics[width=.99\textwidth]{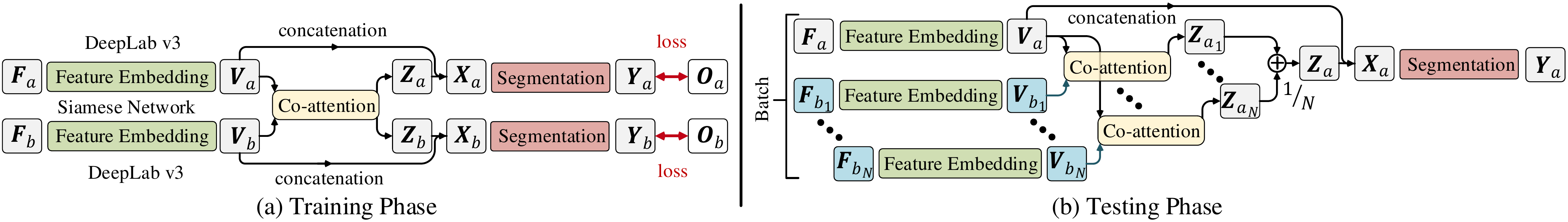}
	\caption{Schematic illustration of training pipeline (a)  and  testing pipeline (b) of COSNet. }
	\label{fig:pipeline}
\vspace*{-8pt}
\end{figure*}

After obtaining the similarity matrix $\mathbf{S}$, as shown in the green and red areas in Fig.~\ref{fig:matrix}, we normalize $\mathbf{S}$ row-wise and column-wise with a softmax function:
\begin{equation}\small
\mathbf{S}^{\text{\texttt{c}}} = \text{softmax}(\mathbf{S}), \ \ \ \ \
\mathbf{S}^{\text{\texttt{r}}} = \text{softmax}(\mathbf{S}^\top\!)~,
\label{normalize}
\end{equation}
where softmax($\cdot$) normalizes each column of the input.
In Eq.~\ref{normalize}, the $i$-th column of $\mathbf{S}^{\text{c}}$ is a vector with length $W\!H$. This vector reflects the relevance of each feature ($1,...,W\!H$) in $\mathbf{V}_a$ to the $i$-th feature in $\mathbf{V}_b$. Next, the attention summaries for the feature embedding $\mathbf{V}_a$  \wrt $\mathbf{V}_b$ can be computed as (see the blue areas in Fig.~\ref{fig:matrix}):
\begin{equation}\small
\begin{aligned}
&\mathbf{Z}_a \!=\!  \mathbf{V}_b \mathbf{S}^{\text{\texttt{c}}} \!=\! \left[\mathbf{Z}^{(1)}_{a} \  \ \mathbf{Z}^{(2)}_{a} \ ... \ \mathbf{Z}^{(i)}_{a}\ ... \  \mathbf{Z}^{\tiny{(W\!H)}}_{a}  \right]    \in  \mathbb{R}^{C\times (W\!H)}, \\
&\mathbf{Z}^{(i)}_{a} \!=\! \mathbf{V}_{\!b} \otimes \mathbf{S}^{{\text{\texttt{c}}}(i)} = \sum\nolimits_{j=1}^{W\!H} \mathbf{V}^{(j)}_{\!b} \cdot \mathbf{s}^{\text{\texttt{c}}}_{ij} \in  \mathbb{R}^{C}, 
\end{aligned}
\label{summary}
\end{equation}
where $\mathbf{Z}^{(i)}_{a}$ denotes the $i$-th column of $\mathbf{Z}_a$,    `$\otimes$' denotes the matrix times vector, $\mathbf{S}^{{\text{\texttt{c}}}(i)}$ is the $i$-th column of  $\mathbf{S}^{\text{\texttt{c}}}$, $\mathbf{V}^{(j)}_{\!b}$ indicates the $j$-th column of $\mathbf{V}^{(j)}$ and $\mathbf{s}^{\text{\texttt{c}}}_{ij}$ is the $j$-th  element in $\mathbf{S}^{{\text{\texttt{c}}}(i)}$.
Similarly, for frame $\mathbf{F}_b$, we compute the corresponding co-attention enhanced feature as: $\mathbf{Z}_b = \mathbf{V}_{\!a} \mathbf{S}^{\text{\texttt{r}}}$.

\noindent\textbf{Gated co-attention.}
Considering the underlying appearance variations between input pairs, occlusions, and background noise, it is better to weight the information from different input frames, instead of treating all the co-attention information equally. To this end, a self-gate mechanism is introduced to allocate a co-attention confidence to each attention summary. The gate is formulated as follows:
\begin{equation}\small
\begin{aligned}
f_g(\mathbf{Z}_{a}) &= \sigma(\mathbf{w}_f \mathbf{Z}_{a} + b_f) \in [0,1]^{W\!H},\\
f_g(\mathbf{Z}_{b}) &= \sigma(\mathbf{w}_f \mathbf{Z}_{b} + b_f) \in [0,1]^{W\!H},
\end{aligned}	
\label{equ:gate}
\end{equation}
where $\sigma$ is the logistic sigmoid activation function, and $\mathbf{w}_f$ and $b_f$ are the convolution kernel and bias, respectively.
The gate $f_g$ determines how much information from the reference frame will be preserved and can be learned automatically. 
After calculating the gate confidences, the attention summaries are updated by:
\begin{equation}\small
\mathbf{Z}_{a} = \mathbf{Z}_{a}\star f_g(\mathbf{Z}_{a}),  \ \ \ \
\mathbf{Z}_{b} = \mathbf{Z}_{b}\star f_g(\mathbf{Z}_{b}),
\label{eq:updatecoattention}	
\end{equation}
where `$\star$' denotes channel-wise Hadamard product. These operations lead to a \textit{gated co-attention} framework.

Then we concatenate the final co-attention representation $\mathbf{Z}$ and the original feature $\mathbf{V}$ together:
\begin{equation}\small
\mathbf{X}_a \!=\! [\mathbf{Z}_a, \mathbf{V}_a] \!\in\! \mathbb{R}^{W\!\times \!H\!\times \!2C},  \ \ \mathbf{X}_{b} \!=\! [\mathbf{Z}_{b}, \mathbf{V}_{b}] \!\in\! \mathbb{R}^{W\!\times\! H\!\times\! 2C},
\label{equ:concat}
\end{equation}
where `[$\cdot$]' denotes the concatenation operation. Finally, the co-attention enhanced feature $\mathbf{X}$ can be fed into a segmentation network to produce a final result $\mathbf{{Y}}\in[0,1]^{W\!\times\! H}$.

\subsection{Full COSNet Architecture}\label{sec:architecture}
Fig.~\ref{fig:pipeline} shows the training and testing pipelines of the proposed COSNet.
Basically, COSNet is a Siamese network which consists of three cascaded parts:  a DeepLabv3~\cite{DBLP:journals/corr/ChenPSA17} based feature embedding module, a co-attention module (detailed in \S\ref{sec:method}) and a segmentation module.

\noindent\textbf{Network architecture during training phase.} 
In the training phase, the Siamese network based COSNet takes two streams as input, \ie, a pair of the frame images $\{\mathbf{F}_{\!a}, \mathbf{F}_{\!b}\}$ which are randomly sampled from the same video. First, the feature embedding module is used to build their feature representations: \!$\{\mathbf{V}_{\!a}, \!\mathbf{V}_{\!b}\}$. Next, $\{\mathbf{V}_{\!a}, \!\mathbf{V}_{\!b}\}$ are refined by the co-attention module and the co-attention enhanced feature $\{\mathbf{X}_a, \mathbf{X}_b\}$ are computed through Eq.~\ref{equ:concat}. Finally, the corresponding segmentation predictions $\{\mathbf{Y}_{\!a}, \mathbf{Y}_{\!b}\}$ are produced by the segmentation module which consists of multiple small kernel convolution layers.  Detailed configurations of the three modules can be found in the next section.

As we discussed in \S\ref{sec:intro}, primary objects in videos have two essential properties: (i) intra-frame discriminability, and (ii) inter-frame consistency. To distinguish the  foreground target(s) from the background (property (i)), we utilize data from existing salient object segmentation datasets~\cite{cheng2015global,DBLP:conf/cvpr/YangZLRY13} to train our backbone feature embedding module. As primary salient object instances are annotated in each image of these datasets, the learned feature embedding can catch and discriminate the objects of most interest. Meanwhile, to ensure COSNet is able to capture the global inter-frame coherence of the primary video objects (property (ii)), we train the whole COSNet with video segmentation data, where the co-attention module plays a key role in capturing the correlations between video frames. Specifically, we take two randomly selected frames in a video sequence to build training pairs. It is worth mentioning that this operation naturally and effectively augments training data, compared to previous recurrent neural network based UVOS models that take only consecutive frames.

In this way, the COSNet is alternatively trained with static image data and dynamic video data. When using image data, we only train the feature embedding module, where an extra $1\!\times\!1$ convolution layer with \textit{sigmoid} activation is added to generate intermediate segmentation side-output. The video data is used to train the whole COSNet, including the feature embedding module, the co-attention module as well as the segmentation module. We employ the weighted binary cross entropy loss to train the network:
\begin{equation}\small
\!\!\!\!\mathcal{L}_{\mathcal{C}}(\mathbf{Y}\!, \!\mathbf{O})\!=\!-\!\!\sum\nolimits_{x} \!(1\!-\!{\eta}) o_{x\!}\log({y_{x}})\! +\!  {\eta}(1\!-\!o_{x})\log(1\!-\!{y_{x}}),\!\!
\label{focal_loss}	
\end{equation}
where $\mathbf{O}\!\in\!\{0,1\}^{W\!\times\! H}$ denotes the binary ground-truth, $y_{x}$ is the intermediate or final segment prediction $\mathbf{Y}$ at pixel $x$, and
$\eta$ is the foreground-background pixel number ratio. 

In addition, for the symmetric co-attention in Eq.~\ref{orthonal}, we add an extra orthogonal regularization into the loss function to maintain the symmetry of weight matrix  $\mathbf{W}$:
\begin{equation}\small
\mathcal{L}\!=\!\mathcal{L}_{\mathcal{C}}+\lambda \left|\mathbf{W}\mathbf{W}^\top \!- \mathit{I}\right|,
\label{orthonal_regular}
\end{equation}
where $\lambda$ is the regularization parameter.

\begin{figure*}[t]
	\centering
	\includegraphics[width=.98 \textwidth]{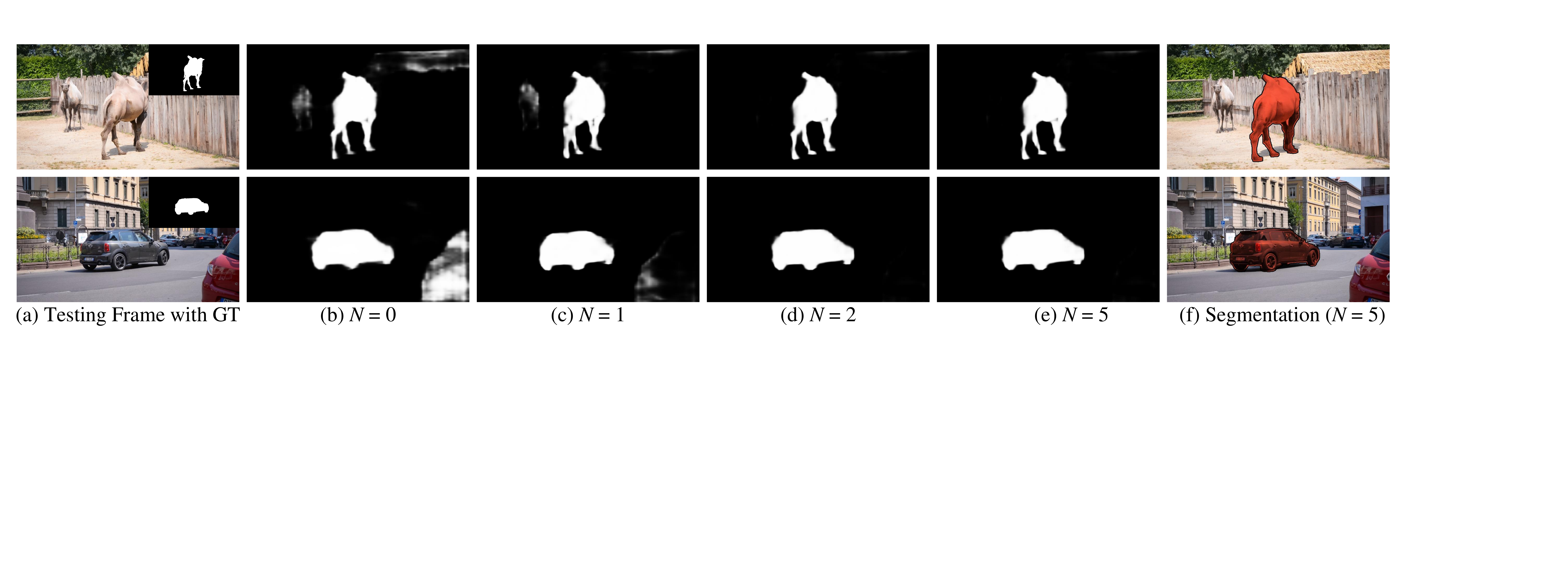}
	\caption{\small Performance improvement for an increasing number of reference frames (\S\ref{sec:ablation}). (a) Testing frames with ground-truths overlaid. (b)-(e) Primary object predictions with considering different number of reference frames ($N\!=\!0,1,2,$ and $5$). (f) Binary segments through applying CRF to (e). We can see that without co-attention, the COSNet degrades to a frame-by-frame segmentation model ((b): $N\!=\!0$). Once co-attention is added ((c): $N\!=\!1$), similar foreground distraction can be suppressed efficiently. Furthermore, more inference frames contribute to better segmentation performance ((c)-(e)). }
	\label{fig:fig3}
\vspace*{-8pt}
\end{figure*}

\noindent\textbf{Network architecture during testing phase.} Once the network is trained, we apply the COSNet to unseen videos. Intuitively, given a test video, we can feed each frame to be segmented, along with only one reference frame sampled from the same video, into the COSNet successively. Performing this operation frame-by-frame, we can obtain all the segmentation results. However, with such a simple strategy, the segmentation results still contain considerable noise, since the rich and global correlation information in the videos is not fully explored. 
%
Therefore, it is critical to include more references during the testing phase (see Fig.~\ref{fig:pipeline} (b)). One intuitive solution is to feed a set of $N$ different reference frames (uniformly sampled from the same video) into the inference branches and average all predictions. A more favored way is that for the query frame $\mathbf{F}_a$, with the reference frame set $\{\mathbf{F}_{b_n}\}_{n=1}^N$ containing $N$ reference frames, Eq.~\ref{eq:updatecoattention} is further reformulated by considering more attention summaries $\{\mathbf{{Z}}_{a_n}\}_{n=1}^N$:
\begin{equation}\small
\mathbf{Z}_{a} \leftarrow \frac{1}{N}\sum\nolimits_{n=1}^{N} \mathbf{Z}_{a_n}\star f_g(\mathbf{Z}_{a_n}).
\label{equ:average}
\end{equation}
In this way, during the testing phase, the co-attention based feature $\mathbf{Z}_{a}$ is able to efficiently capture the foreground information from a global view by considering more reference frames.  Then we feed $\mathbf{Z}_{a}$ into the segmentation module to generate the final output $\mathbf{Y}_a$. Following the widely used protocol~\cite{DBLP:conf/iccv/TokmakovAS17,DBLP:conf/cvpr/TokmakovAS17,Song_2018_ECCV}, we apply CRF as a post-processing step. In \S\ref{sec:ablation}, we will quantitatively demonstrate the performance improvement with the increasing number of reference frames.

\subsection{Implementation Details}
\noindent\textbf{Detailed network architecture.}\label{sec:dna}
The backbone network of our COSNet is DeepLabv3~\cite{DBLP:journals/corr/ChenPSA17}, 
which consists of the first five convolution blocks from ResNet~\cite{DBLP:conf/cvpr/HeZRS16} and an atrous spatial pyramid pooling (ASPP) module~\cite{DBLP:journals/corr/ChenPSA17}.  For the vanilla co-attention module (Eq.~\ref{co-attention1}), we implement the weight matrix $\mathbf{W}$  using a fully connected layer with $512\!\times\!512$ parameters.
In addition, the channel-wise co-attention in Eq.~\ref{channel-attention1} is built on a Squeeze-and-Excitation (SE)-like module~\cite{hu2018senet}. Specifically, the channel weights generated through fully connected layer with  $512$ nodes in one branch are applied to the feature embedding of the other branch~\cite{hu2018senet}. Eq.~\ref{equ:gate} is implemented with $1\!\times\!1$ convolution layer with sigmoid activation function.
The segmentation module consists of two $3\!\times\!3$ convolutional layers (with 256 filters and batch norm ) and a $1\!\times\!1$ convolutional layer (with 1 filter and $sigmoid$ activation) for final segmentation prediction.

\noindent\textbf{Training settings.}
The whole training procedure of our COSNet consists of two alternated steps. When using static data to fine-tune the DeepLabV3 based feature embedding module, we take advantage of image saliency datasets: MSRA10K~\cite{cheng2015global} and DUT~\cite{DBLP:conf/cvpr/YangZLRY13}. In this way, the pixels
belong to the foreground target tend to close to each
other. Meanwhile, we train the whole model with the training videos in DAVIS16~\cite{perazzi2016benchmark}. In this step, two randomly selected frames from the same sequence are fed into COSNet as training pairs. Given the input RGB frame images of size $473\!\times\! 473\! \times\! 3$, the size of the feature embeddings $\mathbf{V}_{\!a}$ and $\mathbf{V}_{\!b}$ are  $(W\!=\!60, H\!=\!60, C\!=\!512)$.
The entire network is trained using the SGD optimizer with an initial learning rate of 2.5$\times10^{-4}$. During training, the batch size is set to 8 and the hyper-parameter  $\lambda$ in Eq.~\ref{orthonal_regular} is set to  $10^{-4}$.  We implement the whole algorithm with Pytorch. All experiments and analyses are conducted on a Nvidia TITAN Xp GPU and an Intel (R) Xeon E5 CPU. TThe overall training time is about 20 hours and a forward pass with one image (batch) takes around 0.18 seconds in the testing phase.
\vspace*{-5pt}
\section{Experiments}\label{experiment}

\subsection{Experimental Setup}\label{sec:expset}

We conduct experiments on the three most famous UVOS datasets: DAVIS16~\cite{perazzi2016benchmark}, FBMS~\cite{DBLP:journals/pami/OchsMB14} and Youtube-Objects~\cite{DBLP:conf/cvpr/PrestLCSF12} datasets.

\noindent\textbf{DAVIS16} is a recent dataset which consists of 50 videos in total (30 videos for training and 20 for testing). 
Per-frame pixel-wise annotations are offered.
For quantitative evaluation, following the standard evaluation protocol from~\cite{perazzi2016benchmark}, we adopt three metrics, namely region similarity $\mathcal{J}$, boundary accuracy $\mathcal{F}$, and time stability $\mathcal{T}$. 

\noindent\textbf{FBMS} is comprised of 59 video sequences. Different from the DAVIS dataset, the ground-truth of FBMS is sparsely labeled (only 720 frames are annotated).
Following the common setting~\cite{DBLP:conf/iccv/TokmakovAS17,DBLP:conf/cvpr/TokmakovAS17,DBLP:conf/cvpr/KohK17,Li_2018_CVPR,Li_2018_ECCV1,Song_2018_ECCV,cheng2017segflow},
we validate the proposed method on the testing split which consists of 30 sequences. The region similarity $\mathcal{J}$ is used for evaluation.

\noindent\textbf{Youtube-Objects} contains 126 video sequences which belong to 10 objects categories  with more than 20,000 frames in total. We use the region similarity $\mathcal{J}$ to measure the segmentation performance.

\vspace*{-1pt}
\begin{table}
	\centering
	\resizebox{0.47\textwidth}{!}{
		\setlength\tabcolsep{2pt}
		\renewcommand\arraystretch{1.0}
	\begin{tabular}{r|cc|cc|cc}
		\toprule[1pt]
		\multirow{2}{*}{Network Variant~~~~~~~~~~} &\multicolumn{2}{c|}{DAVIS}&\multicolumn{2}{c|}{FBMS}&\multicolumn{2}{c}{Youtube-Objects}\\
        &mean $\mathcal{J}$ & $\Delta$$\mathcal{J}$ &mean $\mathcal{J}$ & $\Delta$$\mathcal{J}$ &mean $\mathcal{J}$ & $\Delta$$\mathcal{J}$ \\ \hline
            \multicolumn{7}{c}{Co-attention Mechanism}\\\hline
		Vanilla co-attention (Eq.~\ref{co-attention1})  & 80.0 & -0.5 &75.2& -0.4& 70.3& -0.2\\
		Symmetric co-attention (Eq.~\ref{orthonal})  & 80.5 & -& 75.6 &- &70.5 &- \\
		Channel-wise co-attention (Eq.~\ref{channel-attention1})  & 77.2 & -3.3 & 72.7 & -2.9& 67.5 & -3.0 \\
        \textit{w/o.} Co-attention  & 71.3 & -9.2 & 70.1 & -5.5 & 62.9 & -7.6\\  \hline
        \multicolumn{7}{c}{Fusion Strategy}\\\hline
		Attention summary fusion (Eq.~\ref{equ:average}) & 80.5 & - & 75.6 &- &70.5 &- \\
		 Prediction segmentation fusion  & 79.5 & -1.0 & 74.2 & -1.4 & 69.9 &-0.6 \\ \hline
		  \multicolumn{7}{c}{Frames Selection Strategy}\\\hline
		  Global uniform sampling & 80.53 & - & 75.61 & - & 70.54 &-0.01 \\
		 Global random sampling & 80.52 & -0.01 & 75.54 &-0.02 &70.55 &-\\ 
		 Local consecutive sampling & 80.26 & -0.27 & 75.52 & -0.09 & 70.43 &-0.12 \\
		\bottomrule[1pt]
	\end{tabular}}
\vspace{2pt}	
	\caption{Ablation study (\S\ref{sec:ablation}) of COSNet on DAVIS16~\cite{perazzi2016benchmark}, FBMS~\cite{DBLP:journals/pami/OchsMB14} and Youtube-Objects~\cite{DBLP:conf/cvpr/PrestLCSF12} datasets with different co-attention mechanisms, fusion strategies and sampling strategies.  }
	\label{variants}
\end{table}

\begin{table}
	\centering
	\resizebox{0.4\textwidth}{!}{
		\setlength\tabcolsep{7pt}
		\renewcommand\arraystretch{1.0}
		\begin{tabular}{r|ccccc}
			\toprule[1pt]
            \multirow{2}{*}{Dataset~~~~~~}&\multicolumn{5}{c}{Number of reference frames ($N$)}\\
            \cline{2-6}
			&0 & 1 & 2 & 5 & 7 \\\hline
			DAVIS& 71.3 & 77.6 & 79.7 & 80.5 &80.5\\
            FBMS& 70.2 & 74.8 &  75.3 &75.6 &75.6\\
            Youtube-Objects&62.9  & 67.7 &70.5  &70.5  &70.5 \\
			\bottomrule[1pt]	
		\end{tabular}
}	
\vspace{2pt}	
	\caption{Comparisons with different numbers of reference frames during the testing stage on DAVIS16~\cite{perazzi2016benchmark}, FBMS~\cite{DBLP:journals/pami/OchsMB14} and Youtube-Objects~\cite{DBLP:conf/cvpr/PrestLCSF12} datasets (\S\ref{sec:ablation}). The mean $\mathcal{J}$ is adopted. }
	\label{comparison}
\vspace*{-12pt}
\end{table}

\begin{table*}
	\centering
	\resizebox{0.96\textwidth}{!}{
		\setlength\tabcolsep{8pt}
		\renewcommand\arraystretch{1.0}
	\begin{tabular}{lccccccccccccccc}
		\toprule[1pt]
		 &&TRC  & CVOS& KEY & MSG &  NLC & CUT& FST &  SFL&LMP &FSEG  & LVO & ARP   &  {PDB}&\\
		&\multirow{-2}{*}{Method}&~\cite{DBLP:conf/cvpr/FragkiadakiZS12}  & ~\cite{DBLP:conf/cvpr/TaylorKS15} & ~\cite{lee2011key}& ~\cite{DBLP:conf/iccv/OchsB11}  & ~\cite{DBLP:conf/bmvc/FaktorI14} & ~\cite{cheng2017segflow}& ~\cite{DBLP:conf/iccv/PapazoglouF13} & ~\cite{DBLP:conf/iccv/KeuperAB15} & ~\cite{DBLP:conf/cvpr/TokmakovAS17} & ~\cite{jain2017fusionseg}&  ~\cite{DBLP:conf/iccv/TokmakovAS17} & ~\cite{DBLP:conf/cvpr/KohK17} & ~\cite{Song_2018_ECCV} &\multirow{-2}{*}{\textbf{COSNet}} \\
		\hline
		\rowcolor{mygray}
		\hline
		\multirow{3}{*}{ $\mathcal{J}$ } & Mean    &47.3 &48.2 &  49.8& 53.3& 55.1& 55.2 & 55.8& 67.4& 70.0 & 70.7 & 75.9 &76.2  & 77.2& \textbf{80.5} \\
		& Recall    &49.3 &54.0 & 59.1& 61.6& 55.8 &57.5 &64.9 &81.4 & 85.0 & 83.0&89.1& 91.1  &90.1& \textbf{94.0}  \\
		\rowcolor{mygray}
		& Decay    &8.3 &10.5 & 14.1& 2.4 & 12.6 & 2.2 & \textbf{0.0}&6.2 &1.3 & 1.5 &\textbf{0.0} & 7.0&0.9&\textbf{0.0} \\\hline
		\multirow{3}{*}{ $\mathcal{F}$ } & Mean    &44.1 &44.7 &42.7 & 50.8& 52.3&55.2 &51.1 &66.7 &65.9 & 65.3& 72.1& 70.6  & 74.5& \textbf{79.4} \\
		\rowcolor{mygray}
	$\mathcal{F}$	& Recall  &43.6  & 52.6& 37.5&60.0 & 61.0& 51.9&51.6 &77.1 &79.2 &73.8 &83.4 & 83.5&  84.4& \textbf{90.4}\\	
		& Decay     &12.9 & 11.7&10.6 & 5.1& 11.4 &3.4 &2.9 & 5.1 &2.5&1.8 & 1.3 &7.9 &\textbf{-0.2}& 0.0
		\\\hline \rowcolor{mygray} \hline
		{ $\mathcal{T}$ } & Mean   &39.1 & \textbf{25.0} & 26.9& 30.2& 42.5 &27.7  &36.6 &28.2 & 57.2&32.8 &26.5 &39.3& 29.1& 31.9		\\\bottomrule[1pt]
	\end{tabular}
}
\vspace{2pt}	
\caption{\small Quantitative results on the test set of DAVIS16~\cite{perazzi2016benchmark}\protect\footnotemark[1] (see \S\ref{sec:qqr}), using the region similarity $\mathcal{J}$, boundary accuracy $\mathcal{F}$ and time stability $\mathcal{T}$. We also report the recall and the decay performance over time for both $\mathcal{J}$ and $\mathcal{F}$. The best scores are marked in \textbf{bold}.  }
	\label{davis}
\vspace*{-8pt}
\end{table*}

\subsection{Diagnostic Experiments}\label{sec:ablation}
In this section, we focus on exploration studies to assess the important setups and components of COSNet. 
The experiments were performed on the test sets of DAVIS16~\cite{perazzi2016benchmark} and FBMS~\cite{DBLP:journals/pami/OchsMB14} as well as the whole Youtube-Objects~\cite{DBLP:conf/cvpr/PrestLCSF12}. The  evaluation criterion is mean region similarity ($\mathcal{J}$).

\noindent\textbf{Comparison of different co-attention mechanisms.}
We first study the effect of different co-attention mechanisms in COSNet, \ie, vanilla co-attention (Eq.~\ref{co-attention1}), symmetric co-attention (Eq.~\ref{orthonal}) and channel-wise co-attention (Eq.~\ref{channel-attention1}). In Table~\ref{variants}, both the fully connected method and the symmetric method achieve better performance than the channel attention mechanism. This proves the importance of space transformation in co-attention. Furthermore, compared with vanilla co-attention, we find symmetric co-attention performs slightly better. We attribute this to the orthogonal constraint which reduces feature redundancy  while preserving the norm of the features unchanged.

\noindent\textbf{Effect of co-attention mechanism.}
When excluding the co-attention module and only using the base feature embedding network (DeepLabv3),  we observe a significant performance drop (mean $\mathcal{J}$: 80.5$\rightarrow$71.3 in DAVIS), clearly showing the effectiveness of our strategy, which leverages co-attention mechanism to model UVOS from a global view.

\noindent\textbf{Attention summary fusion \textit{vs} prediction fusion.} In Eq.~\ref{equ:average}, we fuse the information from other reference frames by averaging the corresponding co-attention summaries. To verify its effectiveness, we implement another alternative baseline \textit{Prediction Fusion}: $\mathbf{Y}_{a} =\frac{1}{N}\sum\nolimits_{n=1}^{N} \mathbf{Y}_{a_n}$, \ie, directly average the predictions by considering different reference frames. The results in Table~\ref{variants} demonstrate the superiority of fusion in the feature embedding space.

\noindent\textbf{Comparison of different frame selection strategies.}
To investigate frame selection strategy during the testing phase on the final prediction, we further conduct a series of experiments using different sampling methods. Specifically, we adopt global random sampling, global uniform sampling as well as local consecutive sampling. From Table~\ref{variants}, it can be observed that both global-level sampling strategy achieve approximate performance but better than local sampling method. Meanwhile, local sampling-based results are still superior to the results obtained from the backbone network. 
Overall comparisons further prove the importance of incorporating global context.

\begin{table}
	\centering
	\resizebox{0.48\textwidth}{!}{
		\setlength\tabcolsep{6pt}
		\renewcommand\arraystretch{1.0}
	\begin{tabular}{ccccccccccccccc}
		\toprule[1pt]
		\multicolumn{1}{c}{Method}
		    &NLC~\cite{DBLP:conf/bmvc/FaktorI14} & FST~\cite{DBLP:conf/iccv/PapazoglouF13} &FSEG~\cite{jain2017fusionseg}  &MSTP~\cite{hu2018unsupervised} &ARP~\cite{DBLP:conf/cvpr/KohK17}\\
		   \hline	\rowcolor{mygray}
	Mean $\mathcal{J}$  & 44.5 & 55.5 &68.4&60.8& 59.8\\
\hline	
\hline	
Method& IET~\cite{Li_2018_CVPR}&OBN~\cite{Li_2018_ECCV1}&PDB~\cite{Song_2018_ECCV}&SFL~\cite{cheng2017segflow}& \textbf{COSNet}\\
		   \hline\rowcolor{mygray}	
	Mean $\mathcal{J}$  &71.9&73.9& 74.0 &56.0 &  \textbf{75.6}\\
\bottomrule[1pt]
	\end{tabular}
}
\vspace{1pt}	
\caption{\small Quantitative performance on the test sequences of FBMS~\cite{DBLP:journals/pami/OchsMB14} 
using region similarity (mean $\mathcal{J}$). 
}
\vspace*{-8pt}
	\label{FBMS}
\end{table}

\noindent\textbf{Influence of the number of reference frames.} It is also of interest to assess the influence of the number of reference frames $N$ on the final performance. Table~\ref{comparison} shows the results for this. When $N$ is equal to $0$, this means that there is no co-attention for segmentation.  We observe  a large performance improvement when $N$ changes from 0 to 1, which proves the importance of co-attention. Furthermore, 
when $N$ changes from 2 to 5, the quantitative results show increased performance.  When we further increase
$N$, the final performance does not change obviously. We set the value of $N$ to 5 in the evaluation experiments.

Fig.~\ref{fig:fig3} further visualizes the qualitative segmentation result for an increasing number of inference frames. When $N\!=\!0$, the feature embedding module has learned to discriminate the foreground target from the background. However, when a similar object distractor  appears (\eg., the small camel in the first row, or the red car in the second row), the feature embedding module fails to capture the primary target, since no ground-truth is given. In this case, the proposed co-attention mechanism can refer to  long-range frames and capture the primary object, thus effectively suppressing the similar target distraction.


\begin{figure*}[t!]
	\centering
	\includegraphics[width=.96\textwidth]{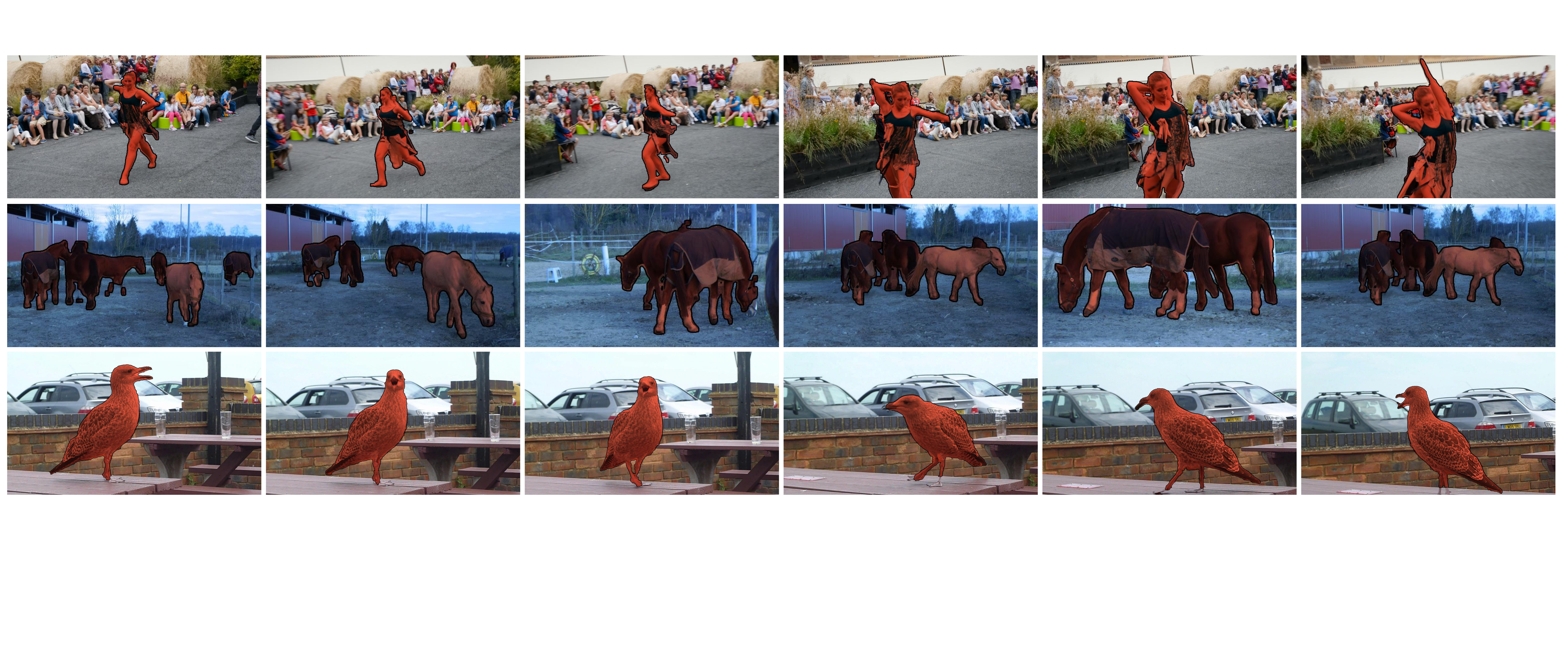}
	\caption{\small Qualitative results on three datasets (\S\ref{sec:qqr}). From top to bottom: \textit{dance-twirl} from the DAVIS16 dataset~\cite{perazzi2016benchmark}, \textit{horses05} from the FBMS dataset~\cite{DBLP:journals/pami/OchsMB14}, and \textit{bird0014} from the Youtube-Objects dataset~\cite{DBLP:conf/cvpr/PrestLCSF12}. }
	\label{qualitative}
\vspace{-6pt}
\end{figure*}
\subsection{Quantitative and Qualitative Results}\label{sec:qqr}
\noindent\textbf{Evaluation on DAVIS16~\cite{perazzi2016benchmark}.} Table~\ref{davis} shows the overall results, with all the top performance methods taken from the DAVIS 2016 benchmark\footnote{\scriptsize{\url{https://davischallenge.org/davis2016/soa_compare.html}}}~\cite{perazzi2016benchmark}. COSNet outperforms all the reported methods across most metrics. Compared with the second best method, PDB~\cite{Song_2018_ECCV}, our COSNet achieves gains of 2.6$\%$ and 4.9$\%$ on $\mathcal{J}$ Mean and  $\mathcal{F}$ Mean, respectively. 

In Table~\ref{davis}, several other deep learning based state-of-the-art UVOS methods~\cite{cheng2017segflow,DBLP:conf/cvpr/TokmakovAS17,jain2017fusionseg,DBLP:conf/iccv/TokmakovAS17,Li_2018_ECCV1}
leverage both appearance as well as extra motion information to improve the performance. 
Different from these methods, the proposed COSNet only utilizes appearance information  but achieves superior performance. We attribute our performance improvement to the consideration of more temporal information through the co-attention mechanism. Compared with these methods using optical flow to catch successive temporal information,  the advantage of exploiting the temporal correlation from a global view is clear when dealing with similar target distractions.

\noindent\textbf{Evaluation on FBMS~\cite{DBLP:journals/pami/OchsMB14}.} We also perform experiments on the FBMS dataset for completeness. Table~\ref{FBMS} shows that our COSNet performs better (75.6\% in mean $\mathcal{J}$) than state-of-the-art methods~\cite{DBLP:conf/bmvc/FaktorI14,DBLP:conf/iccv/PapazoglouF13,jain2017fusionseg,hu2018unsupervised,DBLP:conf/cvpr/KohK17,Li_2018_CVPR,Li_2018_ECCV1,Song_2018_ECCV,cheng2017segflow}.  
In most competing methods, except for the RGB input, additional optical flow information is utilized to estimate the segmentation mask. Considering lots of foreground objects in FBMS share similar appearance with the background but have different motion patterns, optical flow information clearly benefits the prediction.  By contrast, our COSNet only takes advantage of the original RGB information and achieves better performance.

\begin{table}
	\centering
	\resizebox{0.42\textwidth}{!}{
		\setlength\tabcolsep{2pt}
		\renewcommand\arraystretch{1}
	\begin{tabular}{ccccccccc}
		\toprule[1pt]
		& FST & COSEG& ARP & LVO & {PDB}& FSEG & SFL&\\ 	
		\multirow{-2}{*}{Method}&~\cite{DBLP:conf/iccv/PapazoglouF13} &~\cite{tsai2016semantic}& ~\cite{DBLP:conf/cvpr/KohK17} &~\cite{DBLP:conf/iccv/TokmakovAS17} &~\cite{Song_2018_ECCV}&~\cite{jain2017fusionseg} &~\cite{cheng2017segflow}&\multirow{-2}{*}{\textbf{COSNet}}\\ \hline
\rowcolor{mygray}
		\hline
		Airplane (6) & 70.9 &69.3 &73.6 &86.2 &78.0&{81.7}&65.6 & 81.1\\		
		Bird (6) &70.6 &76.0 &56.1 &{81.0} & 80.0& 63.8 & 65.4& 75.7\\
		\rowcolor{mygray}
		Boat (15) & 42.5& 53.5 &57.8 &68.5& 58.9& 72.3& 59.9 &71.3\\
		Car (7) &65.2 &70.4 &33.9 &69.3& 76.5&74.9&64.0 &77.6\\
		\rowcolor{mygray}
		Cat (16) &52.1 &66.8 &30.5 &58.8 &63.0&68.4&58.9 &66.5\\
		Cow (20)&44.5 &49.0 & 41.8&68.5&64.1&68.0 &51.1 & 69.8\\
		\rowcolor{mygray}
		Dog (27) &65.3 &47.5 &36.8&61.7 &70.1&69.4& 54.1 &76.8\\
		Horse (14)& 53.5 &55.7 &44.3 &53.9 &67.6&60.4&64.8 &67.4\\
		\rowcolor{mygray}
		Motorbike (10) &44.2 &39.5 & 48.9&60.8&58.3&62.7 &52.6 &67.7\\
		Train (5) &29.6 &53.4 &39.2 &66.3&35.2&62.2& 34.0 &46.8\\\hline
		\rowcolor{mygray} \hline
		Mean $\mathcal{J}$ &53.8 &58.1 &46.2 &67.5&65.4&{68.4}& 57.0&\textbf{70.5}\\	\bottomrule[1pt]
	\end{tabular}
}
\vspace{2pt}	
\caption{\small Quantitative performance of each category  on Youtube-Objects~\cite{DBLP:conf/cvpr/PrestLCSF12} (\S\ref{sec:qqr}) with the region similarity (mean $\mathcal{J}$). We show the  average performance for each of the 10 categories	from the dataset and the final row shows an average over all the videos.   
}
\vspace*{-10pt}
	\label{Youtube-Objects}
\end{table}

\noindent\textbf{Evaluation on Youtube-Objects~\cite{DBLP:conf/cvpr/PrestLCSF12}.} Table~\ref{Youtube-Objects} illustrates the results of all compared methods for different categories. Our approach outperforms all compared methods~\cite{DBLP:conf/iccv/PapazoglouF13,tsai2016semantic,DBLP:conf/cvpr/KohK17,DBLP:conf/iccv/TokmakovAS17,Song_2018_ECCV,jain2017fusionseg,cheng2017segflow} by a large margin.  FSEG performs second best under the mean $\mathcal{J}$ metric. It is worth noting that the Youtube-Objects dataset shares categories with the training samples in FSEG, which contributes to the enhanced performance~\cite{jain2017fusionseg}. In addition, all the categories in Youtube-Objects can be divided into two types: grid objects (\eg., \textit{Airplane}, \textit{Train}) and non-grid objects (\eg., \textit{Bird}, \textit{Cat}). Despite the objects in the latter class often undergoing shape deformation and quick appearance variation,  the COSNet can capture long-term dependency and handle these scenarios  better than all compared methods. 

\noindent\textbf{Qualitative Results.}
Fig.~\ref{qualitative} shows the qualitative results across three
datasets.  DAVIS16~\cite{perazzi2016benchmark} contains many challenging videos with fast motion, deformation and multiple instances of the same category. We can see that the proposed COSNet can track the primary region or target tightly by leveraging a co-attention scheme to consider global temporal information. The co-attention mechanism helps the proposed COSNet to segment out primary objects from the cluttered background. The effectiveness can also be seen in the \textit{bird0014} sequence of the Youtue-Objects dataset.
In addition, we observe that some videos contain multiple moving targets (\eg., \textit{horses05}) in the FBMS dataset, and the proposed COSNet can deal with such scenarios well.
\vspace*{-10pt}
\section{Conclusion}
\vspace*{-2pt}
By regarding UVOS as a temporal coherence capturing task, we proposed a novel model, COSNet, to estimate the primary target(s).  Through an alternated network training strategy with saliency image and video pairs, the proposed network learns to discriminate primary objects from the background in each frame and capture the temporal correlation across frames. The proposed method achieved superior performance on three representative video segmentation datasets.  Extensive experimental results proved that our method can effectively suppress similar target distraction despite no annotation being given during the segmentation. The COSNet is a general framework for handling sequential data learning, and can be readily extended to other video analysis tasks, such as video saliency detection and optical flow estimation.

\noindent\textbf{Acknowledgements} This work was  supported in part by the National Key Research and Development Program of China (2016YFB1001003), STCSM(18DZ1112300), and the Australian Research Council's Discovery Projects funding
scheme (DP150104645).

{\small
\bibliographystyle{ieee_fullname}
\bibliography{egbib}
}

\end{document}